%% file: egpaper_for_review08-12-overleaf.tex
\documentclass[10pt,twocolumn,letterpaper]{article}
\usepackage{enumitem}
\usepackage{wacv}
\usepackage{times}
\usepackage{epsfig}
\usepackage{graphicx}
\usepackage{rotating}
\usepackage{amsmath}
\usepackage{amssymb}
\usepackage[font=small]{caption}
\usepackage{subcaption}
\usepackage{multirow}
\graphicspath{{./images/}}
\usepackage{tablefootnote}
\setlist[itemize]{leftmargin=*}
\wacvfinalcopy
\usepackage{ragged2e}

\def\FPS3K{FPS3K }

\def\BL{{\sf{BL}}}
\def\RF{{\sf{RF}}}
\def\RP{{\sf{RP}}}
\def\Rp{{\sf{R75}}}
\def\Rh{{\sf{R45}}}
\def\Rf{{\sf{R0}}}
\def\RP{{\sf{RP}}}
\def\REP{{\rm{rep}}}
\def\LMD{{\rm{lmd}}}

\usepackage{array}
\newcolumntype{L}[1]{>{\RaggedRight\hspace{0pt}}m{#1}}
\newcolumntype{R}[1]{>{\RaggedLeft\hspace{0pt}}m{#1}}
\usepackage[pagebackref=true,breaklinks=true,letterpaper=true,colorlinks,bookmarks=false]{hyperref}

\newcommand{\argmin}{\operatornamewithlimits{argmin}}
\ifwacvfinal\fi
\renewcommand{\wrt}{ with respect to }

\begin{document}

\title{ Face Recognition Using Deep Multi-Pose Representations }




\author{
Wael AbdAlmageed$^{a, 0}$
 \quad Yue Wu$^{a, 0}$
 \quad Stephen Rawls$^{a, 0}$
 \quad Shai Harel$^{c}$
 \quad Tal Hassner$^{a, c}$\\
 \quad Iacopo Masi$^{b}$
 \quad Jongmoo Choi$^{b}$
 \quad Jatuporn Toy Leksut$^{b}$
 \quad Jungyeon Kim $^{b}$
 \quad Prem Natarajan$^{a}$\\
 \quad Ram Nevatia$^{b}$
 \quad Gerard Medioni$^{b}$\\
 \and \\
\begin{tabular}{cc}
  $^{a}$Information Sciences Institute & $^{b}$ Institute for Robotics and Intelligent Systems \\
University of Southern California & University of Southern California \\
Marina Del Rey, CA & Los Angeles, California\\
\end{tabular}\\
\\
\begin{tabular}{c}
$^{c}$The Open University\\	
Raanana, Israel
\end{tabular}
}

\maketitle
\ifwacvfinal\thispagestyle{empty}\fi
\maketitle
\ifwacvfinal\thispagestyle{empty}\fi

\begin{abstract}
\input{abstract}
\end{abstract}

\section{Introduction}
\input{introduction}

\begin{figure*}[!htb]
\centering\scriptsize
\includegraphics[width=1\linewidth]{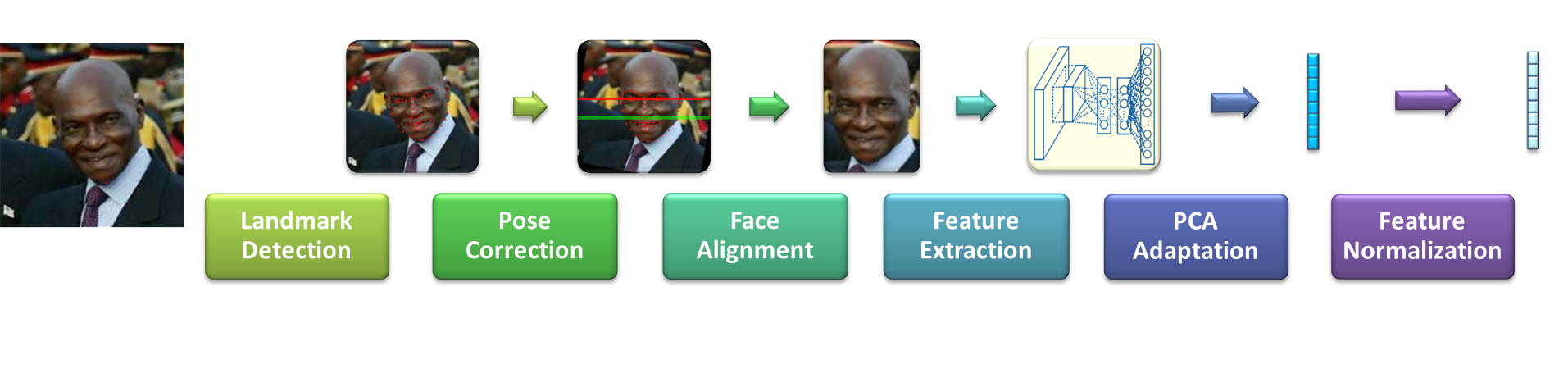}
\caption{Facial representation pipeline.}
\label{fig.pip}
\end{figure*}

\section{Facial Datasets Employed} \label{sec.face_dataset}
\input{data}

\begin{figure}[!h]
\centering\scriptsize
\includegraphics[width=.65\linewidth]{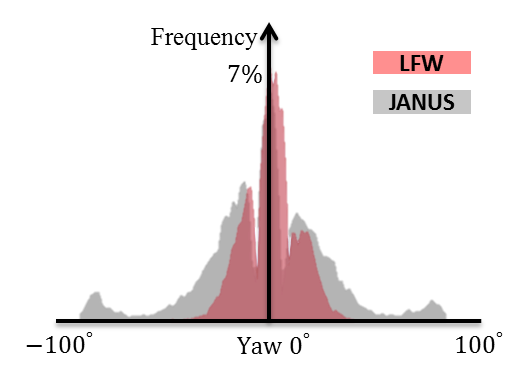}
\caption{Face pose distribution in LFW and JANUS datasets.}\label{fig.dist}
\end{figure}

\section{Face Recognition Pipeline}\label{sec.approach}
\subsection{From Images to Representations}
Given an input face image $X$, a typical recognition pipeline applies a sequence of processing steps in order to transform the 2D color image into a fixed-dimensional feature vector representation, as shown in Figure \ref{fig.pip}. To simplify future discussions, we refer to this transformation process as a function $\REP(\cdot)$, and the resulting feature representation of an image $X$ as $F_X=\REP(X)$. In the following we discuss the processing steps of $\REP(\cdot)$.

\subsubsection{Facial Landmark Detection and Face Alignment}
A facial landmark detection algorithm, $\mathrm{lmd}$, takes a face image $X$ and estimates $n$ predefined key points, such as eye corners, mouth corners, nose tip, \etc, as shown in Equation \eqref{eq.lmd} 
\begin{equation}\label{eq.lmd}
\LMD(X) = \left[\begin{array}{cc}
P^1_x&P^1_y\\
P^2_x&P^2_y\\
\vdots&\vdots\\
P^n_x&P^n_y\\
\end{array}
\right]
\end{equation}
where $P_x$ and $P_y$ denote a key point's coordinate along $x$ and $y$ axes, respectively. Depending on the number of defined facial key points, landmarks  can be roughly classified into two types --- (1) sparse landmark, such as in the 5-point multitask facial landmark dataset \cite{tcdcn} and (2) dense landmark, such as in the 68-point 300w dataset \cite{300wb}. 
Regardless of the facial landmark detector used, landmark detection produces anchor points of a face as a preprocessing step to face recognition and other face analysis tasks.


Detected landmarks are then used to estimate the roll angel of the face pose. The image is then pose-corrected by rotating the image. Since landmark detection algorithms are often trained with imbalanced datasets, which creates some detection errors, the landmark detection is re-applied to the pose-corrected image, as shown in Figure \ref{fig.pip}.

In order for downstream processing steps, such as feature extraction and recognition, to work on consistent feature spaces, all face images must be brought to a common coordinate system through face alignment, which reduces pose variations. This is attained by aligning detected landmarks with a set of landmarks from a reference model using distance minimization. If the reference model is in 2D, the process is called in-plane alignment and if it is a 3D model the process is called out-of-plane alignment.  In our recognition pipeline, we use both non-reflective similarity transformation for in-plane alignment, and {a perspective transformation for out-of-plane alignment}. 

Specifically, given a set of reference facial landmarks $\LMD(R)$, we seek similarity transformation $T$ to align a face image to the reference model, such that 
\begin{equation}
T^*=\argmin_{T}\left\|T \,[\,\LMD(X)\,|\,\mathbf{1}\,]' - [\,\LMD(R)\,|\,\mathbf{1}]\,\right\|_2^2
\end{equation}
where $[\LMD(X)\,|\, 1]$ is simply an expansion of $\LMD(X)$ by adding an all-one vector $\mathbf{1}$, and $T$ is a homogeneous matrix defined by rotation angle $\theta$, scaling factor $s$, and translation vector $[t_x,t_y],$ as shown in Equation\eqref{eq.simi_mat}.
\begin{equation}\label{eq.simi_mat}
T=\begin{bmatrix}
s\cos\theta & -s\sin\theta & t_x\\
s\sin\theta & s\cos\theta & t_y\\
0 & 0 & 1\\
\end{bmatrix}
\end{equation}

\begin{figure*}[!htb]
\centering\scriptsize
\begin{tabular}{@{}c@{}m{.1cm}@{}c@{}m{.1cm}@{}c@{}m{.1cm}@{}c@{}}
\includegraphics[width=.15\linewidth]{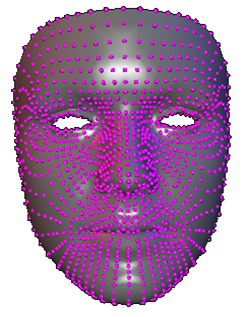}{(a)} &&
\includegraphics[width=.15\linewidth]{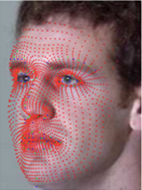}{(b)}&&
\includegraphics[width=.32\linewidth,trim=0.2cm 0.1cm 0.2cm 0cm, clip]{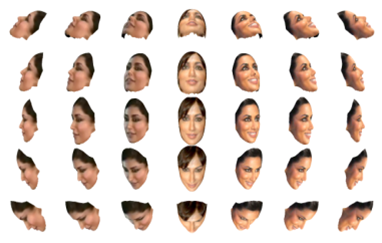}{(c)}&&
\reflectbox{\includegraphics[width=.255\linewidth]{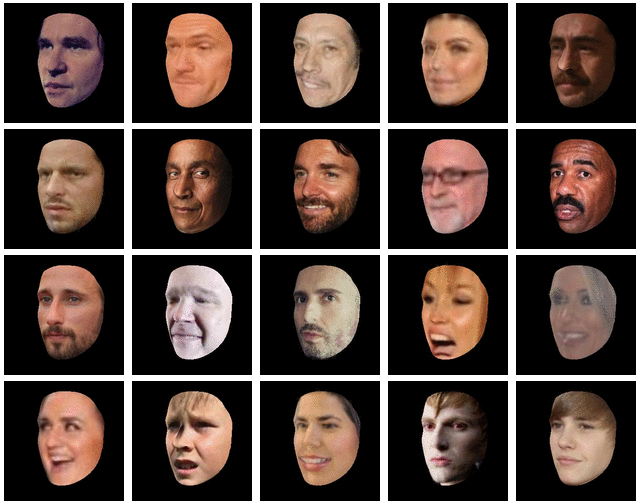}}{(d)}\\
\end{tabular}
\caption{Out-of-plane face alignment via rendering: (a) reference generic 3D face shape; (b) face image with estimated 3D face shape; (c) rendered face at different yaw-and-pitch grids; (d) aligned faces at yaw 45$^\circ$ and pitch-0$^\circ$.}
\label{fig.render}
\end{figure*}

Unlike 2D alignment, our 3D alignment relies on a 3D generic face shape model as shown in Figure \ref{fig.render}, although we still need the detected facial landmark to estimate initial face shape. Once we successfully fit our generic 3D face shape model to a given face image, we can render face images with arbitrary yaw-pitch-roll parameters (details can be found in \cite{render_face} and \cite{talfrontal}). For example, Figure \ref{fig.render}(c) shows rendered face at different yaw and pitch values. Figure \ref{fig.render}(d) shows different face images aligned to the same yaw-pitch-roll configuration.
Finally, aligned images are cropped to a fixed size of {160$\times$128} pixels, which is used in subsequent recognition steps. 

\subsubsection{Feature Extraction and Domain Adaptation}
Feature extraction is an essential module in our representation pipeline, for its responsibility to produce features that provide the power for discriminating between different subjects. Classically, face recognition systems used hand-crafted features.  For example, local binary pattern (LBP) \cite{lbp2006} and its variants \cite{tal1} have been effectively used for face recognition, and Gabor wavelet features have also been widely used \cite{gabor2010}. Since many of such hand-crafted features involve hyper-parameters, multi-scale, multi-resolution, and/or multi-orientation features have also been useful for face recognition, although they require more computations and occupy more space. Often, hand-crafted features do not require learning from data. However, with the availability of training data, higher-level feature representations could be learned \cite{sikka2012exploring}. This learning process still depends on hand-crafted features, rather than raw data. 

In contrast, recent advances in deep learning \cite{msu,umass,umd,multimodal} demonstrate that feature representations from raw data can be learned along with the recognition task, where all layers, except the last one, in a deep neural network (DNN) are considered to be feature learning layers. As a result, one may use DNNs for feature extraction. Details of how we learn feature extraction using DNNs are explained in Section \ref{sec.face.rep}. For now, we deal with a any feature extractor as a black box that takes an aligned and cropped image as as input and produces a high-dimensional feature vector.

The objective of domain adaptation is to close the gap between  training and testing data. We use principal component analysis (PCA) to learn the orthogonal transformation from a testing dataset, such that 95\% of its original feature variance is kept, because it does not require any labeling and is bounded by a low computational complexity. Although feature dimension reduction is not the main purpose here, possible noise is dropped along with the discarded feature dimensions. Although one may directly use adapted features for matching, additional normalization can be very helpful, such as $L_2$  and power normalization \cite{fishervector}. 

\subsection{Face Representation in Practice}\label{sec.face.rep}
Although one may implement the discussed conceptual pipeline in many different ways, we believe that the two most important components of a face recognition pipeline are alignment and feature extraction. Therefore, we mainly focus on the combination of face alignment and face feature extraction, and study the configurations shown in Table \ref{tab.pip}. In this table, HDLBP stands for \textit{high-dimensional local binary pattern} \cite{lbp2006}, and \textit{AlexNet}, \textit{VGG16} and \textit{VGG19} refer to CNN architectures of \cite{krizhevsky2012imagenet}, and config-D and E in \cite{vgg}, respectively; ``Ref. Model'' denotes the reference model used in face alignment, and ``avg-all-face-lmd'' means that we use the averaged landmark vector of all training data, while ``gene-face-yaw@45'' means that we use the generic 3D face model at  $45^{\circ}$ yaw and  $0^\circ$ pitch (see Figure \ref{fig.render}(b)). For HDLBP feature extraction, we compute a patch-based LBP histogram of $6,098$ predefined facial key points, and then concatenate these histograms into a feature vector.

\begin{table}[!h]
\scriptsize\centering
\caption{Face representation pipelines.}\label{tab.pip}
\begin{tabular}{l@{}m{.1cm}@{}r@{}m{.1cm}@{}r@{}m{.1cm}@{}r@{}m{.1cm}@{}r}\hline\hline
\bf{Pip. Acronym} && \bf{Alignment} && \bf{Ref. Model} && \bf{Feature} && \bf{Rep. Dim.}\\
\hline
HLBP && in-plane && avg-face && HDLBP && 100,000 \\\hline
ALEX-AF && in-plane && avg-all-face-lmd && AlexNet && 4,000\\
ALEX-FF && in-plane && avg-frontal-face-lmd&& AlexNet && 4,000\\
ALEX-PF && in-plane && avg-profile-face-lmd&& AlexNet && 4,000\\
ALEX-FY0 && out-of-plane && gene-face-yaw@0 && AlexNet && 4,000\\
ALEX-FY45 && out-of-plane && gene-face-yaw@45 && AlexNet && 4,000\\
\hline
VGG16-AF && in-plane && avg-face-lmd && VGG16 && 4,000\\
\hline
VGG19-AF && in-plane && avg-all-face-lmd && VGG19 && 4,000\\
VGG19-FF && in-plane && avg-frontal-face-lmd && VGG19 && 4,000\\
VGG19-PF && in-plane && avg-profile-face-lmd && VGG19 && 4,000\\
VGG19-FY0 && out-of-plane && gene-face-yaw@0 && VGG19 && 4,000\\
VGG19-FY45 && out-of-plane && gene-face-yaw@45 && VGG19 && 4,000\\
VGG19-FY75 && out-of-plane && gene-face-yaw@75 && VGG19 && 4,000\\
\hline\hline
\end{tabular}
\end{table}

From this point, we concentrate on the discussion of obtaining deep learning features. CASIA-WebFace is used as our standard data set for training and validating different CNN architectures. We preprocessed the dataset in order to obtain aligned and cropped version of the data prior to training the CNNs.

Regardless of the CNN architecture used, we always use CASIA-WebFace as our training and validation dataset. Depending on the  representation pipeline used, we preprocess all CASIA-WebFace data until face alignment. This produces approximately 400,000 samples from 10,500 subjects. 
\subsubsection{Transfer Learning}
Since the amount of face data in CASIA-WebFace is relatively limited ($400,000$ images for $10,500$ subjects), we initialize our CNNs with pre-trained CNNs using the publicly available models from the ILSVRC2014 image classification task, whose ImageNet dataset contains more than 100 million images for 1000 classes. In order to use a pre-trained model as an initial model for the CASIA-WebFace recognition task, we keep all weights of all CNN layers except those from the last dense layer, since the number of output nodes of the last layer must correspond to the number of subject in WebFace (\ie $10,500$) and reinitialize this layer with random weights. After we construct this base model, we begin \emph{transfer learning} process \cite{oquab2014learning} for face recognition in using Caffe library, and obtain new CNN models of ALEX-AF, VGG16-AF, and VGG19-AF, modified to match WebFace subjects and adapted to face feature extraction. Caffe provides pretrained models of AlexNet, VGG16 and VGG19 and automatically performs the required preprocessing, such as resizing an input image to $224\times 224$ pixels, subtracting an average image, \etc. With respect to model training, we use the stochastic gradient descent optimizer with the learning rate starting at $1e-3$ and reducing it to 1e-4 as the convergence plateaus.

\begin{figure}[!h]
\centering\scriptsize
\includegraphics[width=.85\linewidth]{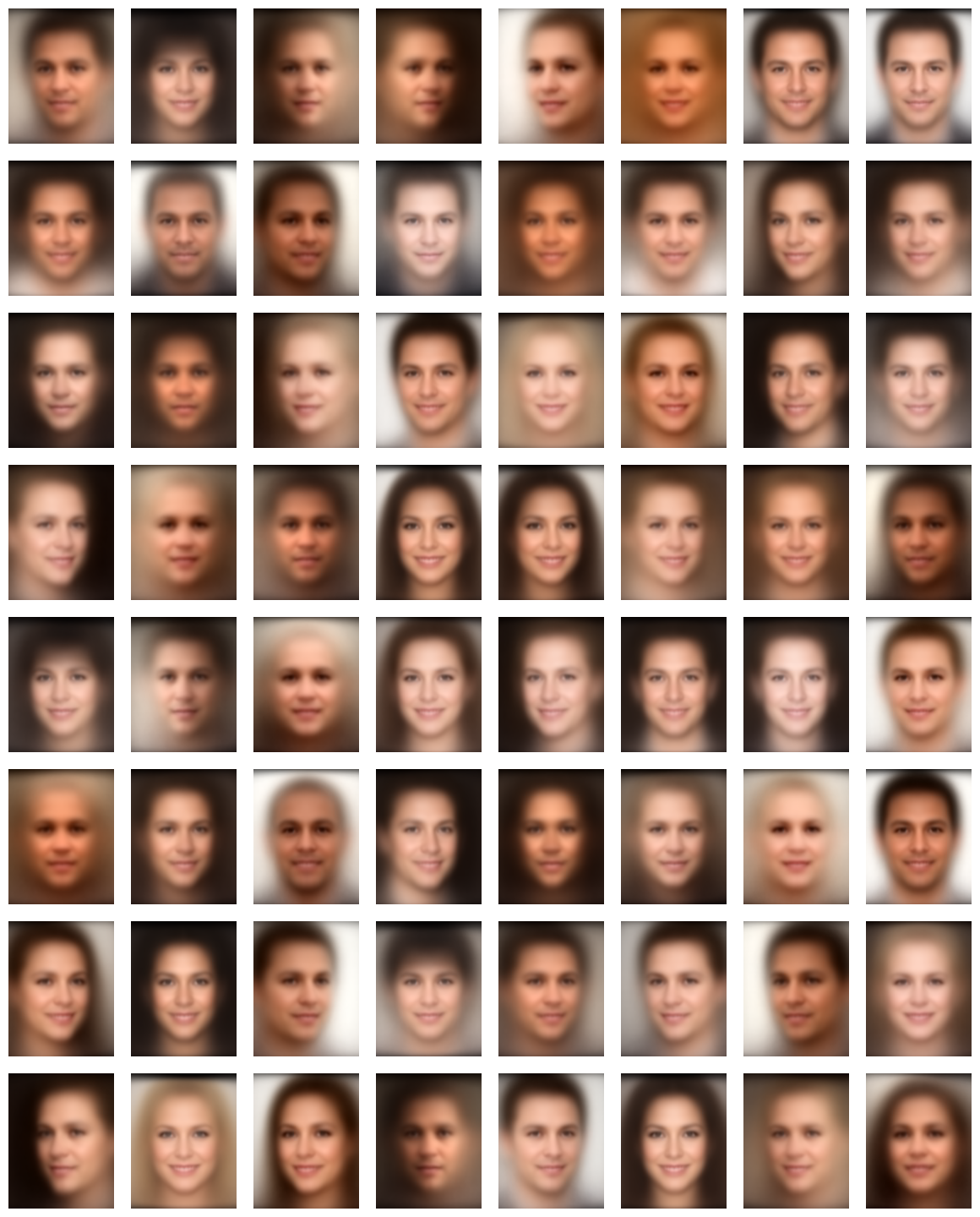}\\{(a)}\\
\includegraphics[width=.85\linewidth]{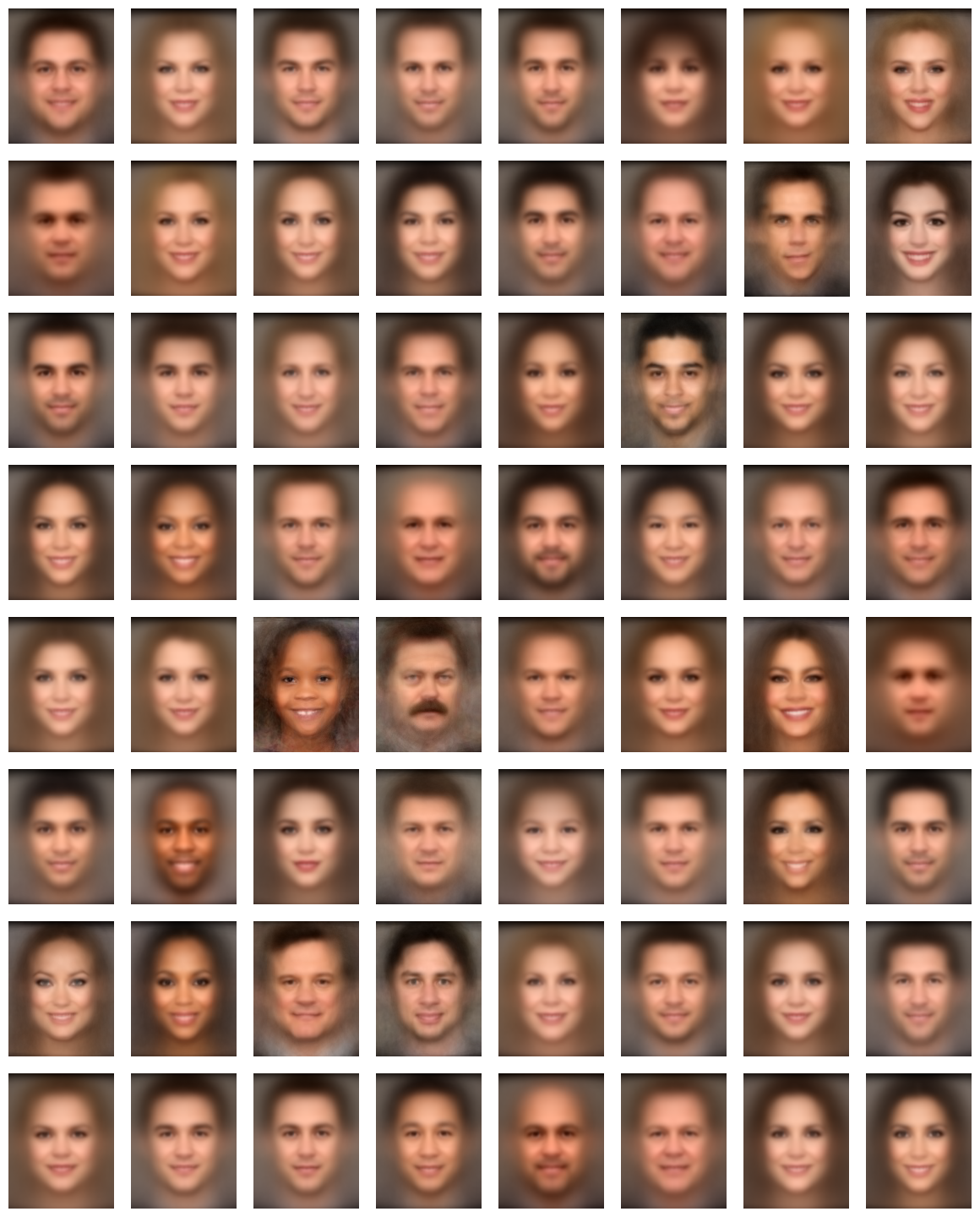}\\{(b)}\\
\caption{GMM 64 Clustering CASIA-WebFace data according to (a) raw RGB-values, and (b) VGG19-AF fc7 feature. }
\label{fig.cnn}
\end{figure}

To verify the performance of transfer learning, we cluster the output of fc7 layer. Figure \ref{fig.cnn} shows cluster-wise average faces from the CASIA-WebFace \cite{casia} data using raw RGB values and the fc7 layer features of VGG19-AF. It can be seen that VGG19-AF learns many important facial attributes, such as gender, face shape, ethnicity, hair color, \etc. 

\subsubsection{Pose-wise Fine Tuning}
\begin{figure}[!h]
\centering\scriptsize
\begin{tabular}{@{}c@{}m{.1cm}@{}c@{}m{.1cm}@{}c@{}m{.1cm}@{}c@{}m{.1cm}@{}c@{}m{.1cm}@{}c@{}}
\includegraphics[width=.16\linewidth]{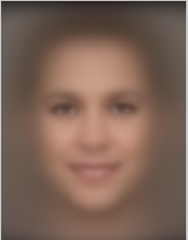}&&
\includegraphics[width=.16\linewidth]{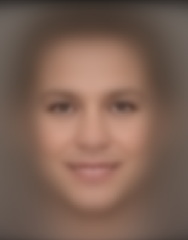}&&
\includegraphics[width=.16\linewidth]{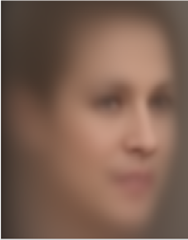}&&
\includegraphics[width=.16\linewidth]{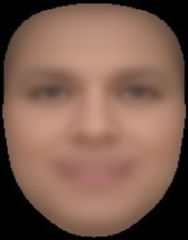}&&
\includegraphics[width=.16\linewidth]{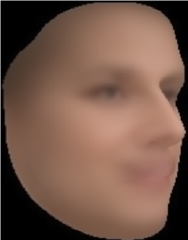}&&
\includegraphics[width=.16\linewidth]{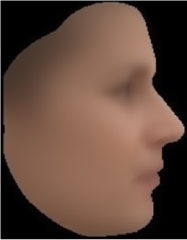}\\
(a)&&(b)&&(c)&&(d)&&(e)&&(f)
\end{tabular}
\caption{Averaged faces of different training partitions. (a) all real, (b) real frontal, (c) real profile, (d) rendered yaw0, (e) rendered yaw45, and (f) rendered yaw75. }
\label{fig.cnn_avg_faces}
\end{figure}

Once we have our face recognition baseline CNN, we then apply \emph{fine tuning} to learn pose-specific features. For example, we use the ALEX-AF model as our base model, and further train ALEX-FF and ALEX-PF, which focus on near-frontal and near-profile faces, respectively. Based on the ALEX-PF CNN, we can further fine-tune it using rendered faces aligned at $0^{\circ}$ yaw, and obtain ALEX-FY0 CNN. The essence behind this CNN training is that we always only move one step forward. For example, we do not use AlexNet as the base model for ALEX-FF because ALEX-Net's training data contains objects from different poses. Similarly, we do not use the near-profile model VGG19-PF as the base model for VGG19-FY0, because the near-frontal CNN model VGG-FF is more appropriate in the sense that rendered faces at $0^{\circ}$ yaw are also frontal. Table 2\label{tab.cnn_learning} gives the full set of descriptions of how we learn these CNNs for face recognition. At the end of each fine tuning process, the last (i.e. classification) layer of the CNN is discarded and the CNN is used as a feature extractor by concatenating the outputs of one or more layers of the CNN. Note that we use different versions of preprocessed CASIA-WebFace data to train different CNNs. Specifically, \textit{all real} refers to all CASIA-WebFace images, \textit{real frontal} and \textit{real profile} only refer to those whose yaw angles are close to $0^{\circ}$ and $75^{\circ}$, respectively. Figure \ref{fig.cnn_avg_faces} shows the averaged images of different training partitions. 

\begin{table}[!h]
\scriptsize\centering
\caption{Deep learning features for face recognition}\label{tab.cnn_learning}
\begin{tabular}{l@{}m{0.2cm}@{}c@{}m{0.2cm}@{}r@{}m{0.2cm}@{}r}\hline\hline
\bf{Pip. Acronym} && \bf{Learning Type} && \bf{Base CNN Model} && \bf{Training Partition}\\\hline
ALEX-AF   && transfer&& AlexNet && all real\\
ALEX-FF   && finetune && ALEX-AF && real frontal\\
ALEX-PF   && finetune && ALEX-AF && real profile\\
ALEX-FY0  && finetune && ALEX-FF && rendered yaw0\\
ALEX-FY45 && finetune && ALEX-PF && rendered yaw45\\\hline
VGG16-AF && transfer && VGG16 && all real\\\hline
VGG19-AF && transfer && VGG19 && all real\\
VGG19-FF   && finetune && VGG19-AF && real frontal\\
VGG19-PF   && finetune && VGG19-AF && real profile\\
VGG19-FY0  && finetune && VGG19-FF && rendered yaw0\\
VGG19-FY45 && finetune && VGG19-PF&& rendered yaw45\\
VGG19-FY75 && finetune && VGG19-FY45&& rendered yaw75\\
\hline\hline
\end{tabular}
\end{table}

\subsection{Multi-Modal Representation for Recognition}
Without loss of generality, assume that we have $k$ distinctive representations of a single input facial image $X$, namely, 
\begin{equation}\label{eq.repR}
R(X)=\{\REP_1(X),\REP_2(X),\cdots,\REP_k(X)\}'
\end{equation}
where each representation is obtained by applying the conceptual pipeline of Figure \ref{fig.pip} using a different feature extractor. When this general multi-modal representation only involves pose-specific models, i.e. ``fine tuned'' models in Table 2, we call this representation a \textbf{multi-pose} representation.

\begin{figure}[!h]
\centering\scriptsize
\includegraphics[width=1.0\linewidth]{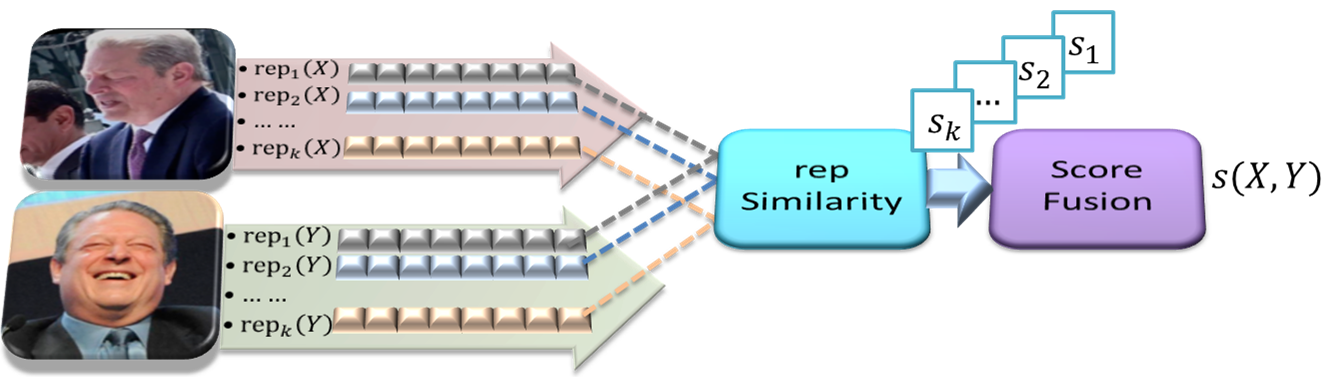}
\caption{Facial matching overview.}
\label{fig.matching}
\end{figure}

Once we define the face representation of Equation \eqref{eq.repR}, we compute the similarity between two face images in two steps --- (1) compare a similarity score between features from the same representation pipeline, and (2) fuse similarities scores across different representation pipelines, as shown in Figure \ref{fig.matching}.
Although one may use various ways to compute feature similarities and to fuse a set of scores, we simply use direct representation-to-representation (i.e. feature-to-feature) comparison in pair-wise fashion to avoid having to construct new classifiers based on the target (i.e. testing) data set. Specifically, we use Equation \eqref{eq.imgsim} to compute the pair-wise image similarity score
\begin{equation}\label{eq.imgsim}
{\rm{sim}}(X,Y) \!\!=\!\! \textrm{fuse}\left( { \{\textrm{rsim}( \REP_j(X), \REP_j(Y)\} |_{j=1}^k } ) \right)
\end{equation}
where we use a cosine similarity metric defined in Equation \eqref{eq.cosine} to quantify the similarity between two features, and use softmax weights to fuse different scores, as shown in Equation \eqref{eq.softmax}, with the bandwidth parameter $\beta=10$.
\begin{equation}\label{eq.cosine}
{\rm{rsim}}(\REP(X),\REP(Y)) = \frac{ <\REP(X),\REP(Y)> }{\|\REP(X)\|\cdot\|\REP(Y)\|}
\end{equation}
\begin{equation}\label{eq.softmax}
\rm{fuse}(\{s_1,s_2,\cdots,s_k\}) = \frac{ \sum_{i=1}^{k} s_i\cdot\exp(\beta\cdot s_k) }{\sum_{i=1}^{k}\exp(\beta\cdot s_k)}
\end{equation}
As mentioned before, since the IJB-A dataset uses the notion of \emph{templates}, we need to compare the similarity between two templates $\mathbb{X}$ and $\mathbb{Y}$ instead of two images. Therefore, we use one more step of softmax fusion over all pairwise scores from images in two templates, as shown in Equation \eqref{eq.tempsim}.
\begin{equation}\label{eq.tempsim}
{\rm{tsim}}(\mathbb{X},\mathbb{Y}) = \textrm{fuse}\left( \{ {\textrm{sim}(X,Y) | X \in\mathbb{X}, Y\in\mathbb{Y} }\}\right)
\end{equation}

\subsection{Time Complexity}
We use a single NVIDIA Tesla K40 GPU for training and testing. Fully training and fine tuning a VGG19-AF-like CNN model takes approximately one week with fully preprocessed images. Testing a single IJB-A/CS2 data split with one CNN model takes roughly one hour using the proposed facial representation pipeline (see Fig. \ref{fig.pip}). The most time consuming step in testing is PCA adaptation and it costs about 20 minutes for a single IJB-A/CS data split.

\section{Experimental Evaluation}\label{sec.eval}

\subsection{Metrics and Protocols}
IJB-A dataset contains two types of protocols, namely \textit{search} and \textit{compare}. The \textit{search} protocol measures the accuracy of open-set and closed-set search among $N$ gallery templates in terms of the true acceptance rate (TAR) at various false acceptance rates (FAR) as well as receiver operating characteristic (ROC) plots. The \textit{compare} protocol measures the verification accuracy between two templates. However, IJB-A carefully designed challenging template pairs by ensuring that subjects of  templates have the same gender, and that their skin colors do not differ more than one level. Metrics of rank-1, rank-5, and the missing rate correspond to false alarm rates of $1 \over 10$ and $1 \over 100$. Detailed descriptions of evaluation   protocols and metrics can be found in \cite{ijba}. 

\subsection{Experiment Overview}
In the rest of this section, we report the results of several experiments based on our proposed face representation and matching scheme. Since each experiment may have its own baseline, comparing performances across different experiments may be inacurate. Although CS2 and IJBA both contains 10 splits of data, we only report averaged scores across these 10 splits to save space. Because CS2 and IJBA provide many associated attributes for a given face image, we do use provided subject face bounding boxes and seed landmarks in most of the following experiments, except those that explicitly state not to. 

\subsection{Selecting CNN Layers for Feature Extraction}
As in many recognition problems, feature plays a core role in face recognition. Given a trained face recognition CNN, we may treat each layer as a feature, and also a collection of features from different layers. Therefore, it is interesting to investigate which combination of features is the best for face recognition. In this experiment, we exhaustively try all possible layer combinations of the last six layers of each CNN architecture, and report face recognition performance on selected  combinations with reasonable performance in Table \ref{tab.feat_comb}, where `x' indicates that this layer of feature is used. 

\begin{table}[!h]
\centering\scriptsize
\caption{DNN Features for Face Recognition in CS2}\label{tab.feat_comb}
\begin{tabular}{@{}r@{}m{.1cm}|@{}m{.1cm}@{}c@{}m{.1cm}@{}c@{}m{.1cm}@{}c@{}m{.1cm}@{}c@{}m{.1cm}@{}c@{}m{.1cm}@{}c@{}m{.1cm}@{}c@{}m{.1cm}@{}c@{}m{.1cm}@{}c@{}m{.1cm}@{}c@{}c@{}m{.1cm}@{}c@{}}
\hline\hline
\bf Layer Name &&& \multicolumn{20}{c}{\bf CNN Feature of Layer Combinations}\\\hline
\bf{prob}   &&& x && x && x && x &&   &&   &&   &&   && x &&   && \\
\bf{fc8}    &&& x &&   &&   &&   && x && x && x && x && x && x && \\
\bf{fc7}    &&&   && x &&   &&   && x &&   && x &&   && x &&   && \\
\bf{fc6}    &&&   &&   && x &&   &&   &&   &&   && x &&   &&   && \\
\bf{pool5}  &&&   &&   &&   && x &&   && x && x && x && x &&   && \\\hline
\bf Metric &&& \multicolumn{20}{c}{\bf ALEX-AF}\\\hline
\bf{TAR@FAR=0.01} &&& .572 && .403 && .374 && .457 && .572 && .687 && .688 && .662 && \bf{.703} && .549\\
\bf{FAR@TAR=0.85} &&& .039 && .058 && .060 && .102 && .042 && .452 && .044 && .041 && \bf{.046} && .038\\
\bf{RANK@10}      &&& .875 && .844 && .820 && .741 && .867 && .885 && .878 && .868 && \bf{.883} && .872\\
\hline
\bf Metric &&& \multicolumn{20}{c}{\bf VGG19-AF}\\\hline
\bf{TAR@FAR=0.01} &&& .816 && .724 && .548 && .749 && .815 && .788 && .789 && .753 && .748 && \bf{.816}\\
\bf{FAR@TAR=0.85} &&& .018 && .018 && .034 && .030 && .017 && .022 && .023 && .028 && .027 && \bf{.017}\\
\bf{RANK@10}      &&& .909 && .892 && .879 && .913 && .909 && .921 && .919 && .916 && .913 && \bf{.909}\\
\hline\hline
\end{tabular}
\end{table}

As one can see, the same collection of features means something different for ALEX-AF and VGG19-AF. Especially, we notice that when we only use the fc7 layer, the second-to-last dense layer, it is the best among all possible layer combinations for VGG19-AF, but not for ALEX-AF. This result clearly confirms that an optimal feature can be a set of features from different layers. From now on, all of our future experiments on ALEX-* representations will be  based on the feature combination of (pool5, fc7, fc8, prob), while VGG* representations are based on the feature combination of (fc8).   

Furthermore, we compare four face recognition pipelines based on different features; specifically HDBLP, ALEX-AF, VGG16-AF, and VGG19-AF as shown in Table \ref{tab.feat}. It is clear that deep learning features outperform classic HDLBP features by a large margin, even though we spend an almost comparable amount of time for grid searching over the HDLBP hyper-parameter space. 

\begin{table}[!h]
\centering\scriptsize
\caption{Feature Influences for Face Recognition in CS2}\label{tab.feat}
\begin{tabular}{@{}r@{}m{.05cm}|@{}m{.05cm}@{}c@{}m{.1cm}@{}c@{}m{.1cm}@{}c@{}m{.1cm}@{}c@{}}
\hline\hline
\bf{Metric} &&& \bf{HDLBP} && \bf{ALEX-AF} && \bf{VGG16-AF} && \bf{VGG19-AF} \\\hline
\bf{TAR@FAR=0.01} &&& .274 && .703 && .779 && .816 \\
\bf{TAR@FAR=0.10} &&& .511 && .906 && .918 && .929 \\
\bf{FAR@TAR=0.85} &&& .680 && .046 && .025 && .017 \\
\bf{FAR@TAR=0.95} &&& .930 && .275 && .228 && .210 \\\hline
\bf{RANK@1}       &&& .398 && .665 && .739 && .773 \\
\bf{RANK@5}       &&& .596 && .834 && .862 && .880 \\
\bf{RANK@10}      &&& .689 && .883 && .895 && .909 \\
\hline\hline
\end{tabular}
\end{table}

\subsubsection{Effect of Landmark Detection Algorithm}
We mainly focus on evaluating facial landmark performance for face recognition. Our baseline of face recognition uses the ALEX-AF representation with cosine similarity and softmax score fusion, and  the dataset used is CS2. Four state-of-the-art facial landmarks are used --- (1) DLIB \cite{dlibpaper} with 68 points, (2) FPS3K \cite{facexpaper} with 68 points, (3) TDCNN \cite{tcdcn} with 5 points, and (4) CLNF with 68 points \cite{clnfpaper1,clnfpaper2} and its variant CLNFs that use seed landmarks provided in CS2 \cite{ijba} for the estimation of an initial face shape. Note that we use DLIB, FPS3K and TDCNN out-of-shelf, and because they do not provide an interface to set the initial facial landmarks, we cannot report their performance with seed landmarks. Detailed results are listed in Table \ref{tab.landmark}. All of these landmark detectors share the same set of face bounding boxes. In general, different landmark detectors do not make a huge performance difference compared to different features, and this implies that CNN features attain a certain level of spatial invariance. On the other hand, initial seed landmarks do help to largely improve face recognition performance.

\begin{table}[!h]
\centering\scriptsize
\caption{Landmark Influence for Face Recognition in CS2}\label{tab.landmark}
\begin{tabular}{@{}r@{}m{.1cm}@{}|@{}m{.1cm}@{}c@{}m{.1cm}@{}c@{}m{.1cm}@{}c@{}m{.1cm}@{}c@{}m{.1cm}@{}c@{}}
\hline\hline
\bf{Metric} &&& \bf{DLIB} && \bf{FPS3K} && \bf{TDCNN} && \bf{CLNF} && {\bf{CLNF}}-s\\\hline
\bf{TAR@FAR=0.01} &&& .689 && .708 && .692 && .682 && .703\\
\bf{TAR@FAR=0.10} &&& .894 && .903 && .906 && .878 && .906\\
\bf{FAR@TAR=0.85} &&& .056 && .049 && .049 && .062 && .046\\
\bf{FAR@TAR=0.95} &&& .249 && .237 && .228 && .504 && .275\\\hline
\bf{RANK@1}       &&& .658 && .636 && .608 && .661 && .665\\
\bf{RANK@5}       &&& .821 && .804 && .803 && .807 && .834\\
\bf{RANK@10}      &&& .871 && .861 && .861 && .862 && .883\\
\hline\hline
\end{tabular}
\end{table}

\input{table6}

\subsection{Single versus Multi-Pose Representations}
We investigate the influence of using three different numbers of representations, namely \textit{single}, \textit{quadruple}, and \textit{quintuple}. Specifically, \textit{single} uses only one of *-AF, *-FF, *-PF, *-FY0, *-FY45, *-FY75 representations, \textit{quadruple} uses the representation tuple of ( *-FF, *-PF, *-FY0, *-FY45 ) and \textit{quintuple} uses the representation tuple of ( *-FF, *-PF, *-FY0, *-FY45, *-FY75 ), where * denotes either ALEX or VGG19 CNN architecture. As shown in Table \ref{tab.representation}, face recognition performance significantly improves as the number of pose representations increases, regardless of whether we use ALEX architecture or VGG19 architecture. 

\subsection{Comparing to State-of-the-Art Methods}
We now compare our VGG19-quintuple representation, namely (VGG19-FF, -PF, -FY0, -FY45, -FY75), to state-of-the-art methods \cite{msu,umass,umd}, along with the baseline GOTS and COTS from \cite{ijba} on the IJB-A dataset. It is clear that our multi-pose representation-based recognition pipeline outperforms other state-of-the-art methods. It is worthy to mention that the algorithm presented in \cite{umd} involves both fine tuning on IJB-A data and uses metric learning using labeled IJB-A training data, while our algorithm is data-agnostic and is used out of the box on both CS2 and IJB-A without target domain specific tuning. 

\input{table78}

\section{Conclusion}\label{sec.conclusion}
We introduced a multi-pose representation for face recognition, which is a collection of face representations learned from specific face poses. We show that this novel representation significantly improves face recognition performance on IJB-A benchmark compared not only to the single best CNN representations but also those state-of-the-art methods that heavily rely on supervised learning, such gallery fine-tuning and metric learning.

\section*{Acknowledgment}
This research is mainly based upon work supported by the Office of the Director of National Intelligence (ODNI), Intelligence Advanced Research Projects Activity (IARPA), via IARPA's 2014-14071600011. The views and conclusions contained herein are those of the authors and should not be interpreted as necessarily representing the official policies or endorsements, either expressed or implied, of ODNI, IARPA, or the U.S. Government.  The U.S. Government is authorized to reproduce and distribute reprints for Governmental purpose notwithstanding any copyright annotation thereon. We thank the NVIDIA Corporation for the donation of the Tesla K40 GPU. 

{\small
\bibliographystyle{ieee}
\bibliography{egbib}
}
\end{document}

%% file: abstract.tex
We introduce our method and system for face recognition using  multiple pose-aware deep learning models. In our representation, a face image is processed by several pose-specific deep convolutional neural network (CNN) models to generate multiple pose-specific features. 3D rendering is used to generate multiple face poses from the input image. Sensitivity of the recognition system to pose variations is reduced since we use an ensemble of pose-specific CNN features. The paper presents extensive experimental results on the effect of landmark detection, CNN layer selection and pose model selection on the performance of the recognition pipeline. Our novel representation achieves better results than the state-of-the-art on IARPA's CS2 and NIST's IJB-A in both verification and identification (i.e. search) tasks. 

%% file: introduction.tex
\footnotetext{Equal contributors and corresponding authors at wamageed, yue\_wu, srawls@isi.edu}
Face recognition has been one of the most challenging and attractive areas of computer vision. The goal of face recognition algorithms is to answer the question, \textit{who is this person in a given image or video frame?} Face recognition algorithms generally try to address two problems --- identify verification and subject identification. Face verification, as known as the $1:1$ matching problem \cite{ijba}, answers the question, \textit{are these two people actually the same?}, while face identification, also known as the $1:N$ problem \cite{ijba}, answers the question, \textit{who is this person, given a database of faces?} 

Labeled Faces in the Wild (LFW) dataset \cite{lfw} is considered one of the most important benchmarks for face recognition research. Recent advances, especially in applying convolutional neural networks (CNN) to face recognition, enabled researchers of achieving close to $100\%$ recognition rates \cite{multimodal}. However, face recognition problem is far from solved, especially in an uncontrolled environment with extreme pose, illumination, expression and age variations. Indeed, as discussed in \cite{ijba}, state-of-the-art commercial and open-source face recognition systems performed far less than satisfactory on the recently released National Institute of Standards and Technology's (NIST) IARPA Janus Benchmark-A (IJB-A), which is considered  much more challenging, than LFW, in terms of the variability in pose, illumination, expression, aging, resolution, \etc \cite{msu,umass,umd}.

IJB-A dataset has been quickly adopted by the research community. In \cite{msu}, the authors represent a face image as a feature vector using a trained convolutional neural network and hash this feature vector to achieve fast face search. Chowdhury et al. in \cite{umass} fine-tunes a trained base-model of a symmetric bilinear convolutional neural network (BCNN) to extract face features, and trains subject-based SVM classifiers to identify individuals. In \cite{umd}, Patel et al. use a trained CNN model to represent a face, and additional joint-Bayesian metric learning to assess the similarity between two face representations.

In this paper we present our face recognition pipeline using a novel multi-pose deep face representation. Unlike previous efforts \cite{umass,umd,msu} that consider pose variations implicitly, we explicitly leverage the variations of face poses by (1) representing a face with different (aligned and rendered) poses using different pose-specific CNNs and (2) perform face similarity comparisons only using same pose CNNs. To our best knowledge, this is the first attempt to use multi-pose in face recognition. This novel approach is applied to IJB-A dataset and is shown to surpasses the state-of-the-art algorithms \cite{umass,umd,msu}, without additional domain adaptation or metric learning.

The remainder of this paper is organized as follows. Section \ref{sec.face_dataset} discusses the training and testing datasets. In Section \ref{sec.approach} we discuss our conceptual single representation along with instances of different pose experts, and the proposed multi-pose representation for face recognition. Evaluation protocols and results of the proposed recognition pipeline are presented in Section \ref{sec.eval}. We conclude the paper in Section \ref{sec.conclusion}.

%% file: data.tex
We use CASIA-WebFace \cite{casia} for training and both IJB-A \cite{ijba} and IARPA's Janus CS2 for evaluation. Following is a brief description of these datasets.

CASIA-WebFace \cite{casia} dataset is the largest known public dataset for face recognition. Specifically, CASIA-WebFace contains $10,575$ subjects with a total of $494,414$ images. We perform the following data clean up steps before using WebFace for training our pose-specific CNN models --- (1) exclude images of all subjects in WebFace that overlap with the testing and evaluation data sets, (2) remove all images of subject with fewer than five images in the dataset and (3) remove images with undetectable faces. Approximately $400,000$ images for $10,500$ subjects remain of which we use $90\%$ and $10\%$ for training and validation, respectively, of the CNNs.

The objective of IARPA's Janus program is push the frontiers of unconstrained face recognition. Janus datasets contains images that have full-pose variations, as illustrated by the pose distribution histogram in Figure \ref{fig.dist}, and provide manual annotations, including facial bounding box, seed landmarks of two-eye and nose base, lighting conditions and some subject attributes, such as ethnicity, gender, \etc. These datasets introduce the novel concept of a \textit{template}, which is a collection of images and videos of the same subject. IJB-A and CS2 share $90\%$ of image data. However, they differ in evaluation protocols. In particular, IJB-A includes protocols for both open-set face identification (i.e. $1:N$ search) and face verification (i.e. $1:1$ comparison), while CS2 focuses on closed-set identification. The datasets are divided into gallery and probe subsets. The gallery subset is a collection of templates for subjects to be enrolled in the system, while the probe subset is a collection of templates for unknown subjects for testing and evaluation purposes.

%% file: table6.tex
\begin{table}[!h]
\centering\scriptsize
\caption{Representation Influences for Face Recognition on CS2}\label{tab.representation}
\begin{tabular}{@{}r@{}m{.1cm}@{}|@{}m{.1cm}@{}c@{}m{.1cm}@{}c@{}m{.1cm}@{}c@{}m{.1cm}@{}c@{}m{.1cm}@{}c@{}m{.1cm}@{}c@{}m{.1cm}@{}c@{}m{.1cm}@{}c@{}}
\hline\hline
\bf{Metric} &&& \bf{AF} && \bf{-FF} && \bf{-PF} && \bf{-FY0} && {\bf{-FY45}} && {\bf{-FY75}} && {\bf{quadruple}} && {\bf{quintuple}}\\\hline
&&&\multicolumn{15}{c}{\bf ALEX- Arch.}\\
\hline
\bf{TAR@FAR=0.01} &&& .703 && .635 && .332 && .754 && .797 && .802 && .814 && .814 \\
\bf{TAR@FAR=0.10} &&& .906 && .729 && .443 && .913 && .935 && .937 && .939 && .941 \\
\bf{FAR@TAR=0.85} &&& .046 && .376 && .558 && .036 && .020 && .019 && .017 && .017 \\
\bf{FAR@TAR=0.95} &&& .275 && .485 && .661 && .249 && .151 && .149 && .143 && .126 \\\hline
\bf{RANK@1}       &&& .665 && .576 && .263 && .706 && .751 && .753 && .781 && .799 \\
\bf{RANK@5}       &&& .834 && .694 && .367 && .844 && .883 && .882 && .898 && .906 \\
\bf{RANK@10}      &&& .883 && .723 && .419 && .889 && .918 && .920 && .928 && .936 \\
\hline
&&&\multicolumn{15}{c}{\bf VGG19- Arch.}\\
\hline
\bf{TAR@FAR=0.01} &&& .816 && .688 && .364 && .806 && .858 && .860 && .873 && .897 \\
\bf{TAR@FAR=0.10} &&& .929 && .738 && .453 && .930 && .948 && .948 && .950 && .959 \\
\bf{FAR@TAR=0.85} &&& .017 && .489 && .659 && .020 && .009 && .009 && .006 && .003 \\
\bf{FAR@TAR=0.95} &&& .210 && .616 && .763 && .171 && .112 && .111 && .101 && .065 \\\hline
\bf{RANK@1}       &&& .733 && .655 && .305 && .769 && .817 && .802 && .854 && .865 \\
\bf{RANK@5}       &&& .880 && .723 && .397 && .881 && .914 && .913 && .927 && .934 \\
\bf{RANK@10}      &&& .909 && .743 && .439 && .912 && .936 && .936 && .946 && .949 \\
\hline\hline
\end{tabular}
\end{table}

\ifx
\begin{table}[!h]
\centering\scriptsize
\caption{Representation Influences for Face Recognition on IJBA}\label{tab.representation}
\begin{tabular}{@{}r@{}m{.1cm}@{}|@{}m{.1cm}@{}c@{}m{.1cm}@{}c@{}m{.1cm}@{}c@{}m{.1cm}@{}c@{}m{.1cm}@{}c@{}m{.1cm}@{}c@{}m{.1cm}@{}c@{}m{.1cm}@{}c@{}}
\hline\hline
\bf{Metric} &&& \bf{AF} && \bf{-FF} && \bf{-PF} && \bf{-FY0} && {\bf{-FY45}} && {\bf{-FY75}} && {\bf{quadruple}} && {\bf{quintuple}}\\\hline
&&&\multicolumn{15}{c}{\bf ALEX- Arch.}\\
\hline
\bf{TAR@FAR=0.01} &&& . && .615 && .325 && .645 && .763 && .797 && . && .788 \\
\bf{TAR@FAR=0.10} &&& . && .725 && .999 && .728 && .929 && .936 && . && .927 \\
\bf{FAR@TAR=0.85} &&& . && .999 && .999 && .666 && .027 && .020 && . && .024 \\
\bf{FAR@TAR=0.95} &&& . && .999 && .999 && .999 && .487 && .389 && . && .184 \\\hline
\bf{RANK@1}       &&& . && .587 && .279 && .616 && .763 && .775 && . && .787 \\
\bf{RANK@5}       &&& . && .703 && .388 && .709 && .897 && .902 && . && .902 \\
\bf{RANK@10}      &&& . && .739 && .448 && .744 && .931 && .935 && . && .933 \\
\hline
&&&\multicolumn{15}{c}{\bf VGG19- Arch.}\\
\hline
\bf{TAR@FAR=0.01} &&& . && . && . && . && . && . && . && . \\
\bf{TAR@FAR=0.10} &&& . && . && . && . && . && . && . && . \\
\bf{FAR@TAR=0.85} &&& . && . && . && . && . && . && . && . \\
\bf{FAR@TAR=0.95} &&& . && . && . && . && . && . && . && . \\\hline
\bf{RANK@1}       &&& . && . && . && . && . && . && . && . \\
\bf{RANK@5}       &&& . && . && . && . && . && . && . && . \\
\bf{RANK@10}      &&& . && . && . && . && . && . && . && . \\
\hline\hline
\end{tabular}
\end{table}
\fi

%% file: table78.tex
\begin{table}[!h]
\centering\scriptsize
\caption{Results on CS2}\label{tab.representation}
\begin{tabular}{@{}r@{}m{.1cm}@{}|@{}m{.1cm}@{}c@{}m{.1cm}@{}c@{}m{.1cm}@{}c@{}m{.1cm}@{}c@{}m{.1cm}@{}c@{}m{.1cm}@{}|c@{}}
\hline\hline
\bf{Metric} &&& \bf{COTS} && \bf{GOTS} && \bf{FV\cite{umd}} && \bf{DCNN-all\cite{umd}} && {\cite{msu}} &&  {\bf{Ours}}\\\hline
\bf{TAR@FAR=0.01} &&& .581 && .467 && .411 && .876 && .733 && .897 \\
\bf{TAR@FAR=0.10} &&& .761 && .675 && .704 && .973 && .895 && .959 \\\hline
\bf{RANK@1}       &&& .551 && .413 && .381 && .838 && .820 && .865 \\
\bf{RANK@5}       &&& .694 && .517 && .559 && .924 && .929 && .934 \\
\bf{RANK@10}      &&& .741 && .624 && .637 && .949 && -    && .949 \\
\hline\hline
\end{tabular}
\end{table}

\begin{table}[!h]
\centering\scriptsize
\caption{1:N Results on IJB-A}\label{tab.representation}
\begin{tabular}{@{}r@{}m{.1cm}@{}|@{}m{.1cm}@{}c@{}m{.1cm}@{}c@{}m{.1cm}@{}c@{}m{.1cm}@{}c@{}m{.1cm}@{}c@{}m{.1cm}@{}|c@{}}
\hline\hline
\bf{Metric} &&& \bf{COTS} && \bf{GOTS} && \cite{umass} && \bf{DCNN-all\cite{umd}} && {\cite{msu}} && {\bf{Ours}}\\\hline
 &&& \multicolumn{10}{c|}{1:N (Search Protocol)}\\\hline
\bf{TAR@FAR=0.01} &&& .406 && .236 && - && - && .733 && .876 \\
\bf{TAR@FAR=0.10} &&& .627 && .433 && - && - && .895 && .954 \\\hline
\bf{RANK@1}       &&& .443 && .246 && .588 && {.860} && .820 && .846 \\
\bf{RANK@5}       &&& .595 && .375 && .797 && {.943} && .929 && .927 \\
\bf{RANK@10}      &&& - && - && - &&  - && -    && .947 \\\hline
 &&& \multicolumn{10}{c|}{1:1 (Verification Protocol)}\\\hline
\bf{TAR@FAR=0.01} &&& - && - && - && - && - && .787 \\
\bf{TAR@FAR=0.10} &&& - && - && - && - && - && .911 \\
\hline\hline
\end{tabular}
\end{table}